\documentclass[10pt,twocolumn,letterpaper]{article}

\usepackage{cvpr}
\usepackage{times}
\usepackage{epsfig}
\usepackage{graphicx}
\usepackage{amsmath}
\usepackage{amssymb}

\usepackage{url}                  % simple URL typesetting
\usepackage{booktabs}             % professional-quality tables
\usepackage{amsfonts}             % blackboard maths fonts
\usepackage{amssymb}              % blackboard maths symbols
\usepackage{mathtools}            % some extra maths tools
\usepackage{mleftright}           % better brackets
\usepackage{graphicx}             % images
\usepackage{tikz}                 % better graphics
\usepackage{pgfplots}             % plotting for tikz
\usepackage{paralist}             % better lists
\usepackage{dcolumn}
\usepackage{xcolor}

\graphicspath{{images/},{graphics/}}
\usetikzlibrary{graphs, shapes, shapes.symbols}

\newcommand{\defoccur}[1]{\textsl{#1}}

\makeatletter
\renewcommand{\paragraph}{%
  \@startsection{paragraph}{4}%
  {\z@}{2ex \@plus 1ex \@minus .2ex}{-1em}%
  {\normalfont\normalsize\bfseries}%
}
\makeatother

\definecolor{blackL}{HTML}{34495E}      % #34495E -- wet asphalt
\definecolor{blackD}{HTML}{2C3E50}      % #2C3E50 -- midnight blue
\definecolor{black}{HTML}{2C3E50}       % #222222 -- black [added]
\definecolor{white}{HTML}{ECF0F1}       % #ECF0F1 -- clouds
\definecolor{whiteD}{HTML}{BDC3C7}      % #BDC3C7 -- silver
\definecolor{grey}{HTML}{95A5A6}        % #95A5A6 -- concrete
\definecolor{greyD}{HTML}{7F8C8D}       % #7F8C8D -- asbestos
\definecolor{blue}{HTML}{3498DB}        % #3498DB -- peter river
\definecolor{blueD}{HTML}{2980B9}       % #2980B9 -- belize hole
\definecolor{blueGreen}{HTML}{1ABC9C}   % #1ABC9C -- turquoise
\definecolor{blueGreenD}{HTML}{16A085}  % #16A085 -- green sea
\definecolor{green}{HTML}{2ECC71}       % #2ECC71 -- emerald
\definecolor{greenD}{HTML}{27AE60}      % #27AE60 -- nephritis
\definecolor{purple}{HTML}{9B59B6}      % #9B59B6 -- amethyst
\definecolor{purpleD}{HTML}{8E44AD}     % #8E44AD -- wisteria
\definecolor{yellow}{HTML}{F1C40F}      % #F1C40F -- sunflower
\definecolor{yellowD}{HTML}{F39C12}     % #F39C12 -- orange
\definecolor{orange}{HTML}{E67E22}      % #E67E22 -- carrot
\definecolor{orangeD}{HTML}{D35400}     % #D35400 -- pumpkin
\definecolor{red}{HTML}{E74C3C}         % #E74C3C -- alizarin
\definecolor{redD}{HTML}{C0392B}        % #C0392B -- pomegranate

\definecolor{dblack}{HTML}{000000}       % #000000 -- black [added]
\definecolor{dwhite}{HTML}{FFFFFF}       % #FFFFFF -- clouds

\newcommand{\reward}{\ensuremath{\mathcal{R}}}

\newcommand\ignore[1]{}
\usepackage{algorithm}
\usepackage{algpseudocode}
\algloopdefx{NoEndIf}[1]{\textbf{if} #1 \textbf{then}}

\def\etal{\textit{et al.}~}

\def\ie{\emph{i.e.},~}
\def\eg{\emph{e.g.},~}
\def\cf{\emph{cf.},~}

\def\etc{\emph{etc.}}

\def \path{\bp C}

\newcommand{\bft}{{\mathbf{t}}}

\newcommand{\bbR}{{\mathbb{R}}}

\usepackage[pagebackref=true,breaklinks=true,letterpaper=true,colorlinks,bookmarks=false]{hyperref}

\usepackage[capitalize, nameinlink]{cleveref} % simpler ref-ing

\cvprfinalcopy % *** Uncomment this line for the final submission

\def\cvprPaperID{1963} % *** Enter the CVPR Paper ID here

\ifcvprfinal\fi
\begin{document}

%%%%%%%%% TITLE
\title{Playing Doom with SLAM-Augmented Deep Reinforcement Learning}

\author{
\begin{tabular*}{0.75\linewidth}{@{\extracolsep{\fill}}ccc}
	Shehroze Bhatti & Alban Desmaison & Ondrej Miksik \\
	Nantas Nardelli & N. Siddharth & Philip H. S. Torr \\
	\multicolumn{3}{c}{University of Oxford}
\end{tabular*}
}

\maketitle
%\thispagestyle{empty}

%%%%%%%%% ABSTRACT
\begin{abstract}
  A number of recent approaches to policy learning in 2D game domains have been
  successful going directly from raw input images to actions.
  However when employed in complex 3D environments, they typically suffer from
  challenges related to partial observability, combinatorial exploration spaces,
  path planning, and a scarcity of rewarding scenarios.
  Inspired from prior work in human cognition that indicates how humans employ a
  variety of semantic concepts and abstractions (object categories,
  localisation, \etc) to reason about the world, we build an agent-model that
  incorporates such abstractions into its policy-learning framework.
  We augment the raw image input to a Deep Q-Learning Network (DQN), by adding
  details of objects and structural elements encountered, along with the agent's
  localisation.
  The different components are automatically extracted and composed into a
  topological representation using on-the-fly object detection and 3D-scene
  reconstruction.
  We evaluate the efficacy of our approach in ``Doom'', a 3D first-person combat
  game that exhibits a number of challenges discussed, and show that our
  augmented framework consistently learns better, more effective policies.
\end{abstract}

%%% Local Variables:
%%% mode: latex
%%% TeX-master: "vizdoom"
%%% End:

%%%%%%%%% BODY TEXT
\vspace{-.25cm}
\section{Introduction}
\label{sec:introduction}

% - What are approaches to learning actions
%
Recent approaches to policy learning in games
\cite{mnih2015human,mnih-atari-2013} have shown great promise and success over a
number of different scenarios.
A particular feature of such approaches is the ability to take the visual game
state directly as input and learn a mapping to actions such that the agent 
effectively explores the world and solves predetermined tasks.
Their success has largely been made possible thanks to the ability of
\defoccur{deep reinforcement learning} (deepRL) networks, neural networks acting
as function approximators within the reinforcement-learning framework.
A particular variant, \defoccur{Deep Q-Learning Networks} (DQN), has been widely
used in a range of different settings with excellent results.
It employs \defoccur{convolutional neural networks} (CNN) as a building block to
effectively extract features from the observed input images, subsequently
learning policies using these features.

For the majority of scenarios that have been tackled thus far, a common
characteristic has been that the domain is 2-dimensional.
Here, going directly from input image pixels to learned policy works well due
to two important factors:
\begin{inparaenum}[i)]
\item a reasonable amount of the game's state is directly observable in the
  image, and
\item a combination of a lower-dimensional action space and smaller exploration
  requirements result in a smaller search space.
\end{inparaenum}
The former ensures that the feature extraction always has sufficient information
to influence policy learning, and the latter makes learning consistent features
easier.
Despite stellar success in the 2D domain, these models struggle in more
complicated domains such as 3D games.

\begin{figure}[!t]
  \centering
  \begin{tabular}{@{}c@{\hspace*{1ex}}c@{\hspace*{1ex}}c@{}}
    \includegraphics[width=0.32\linewidth]{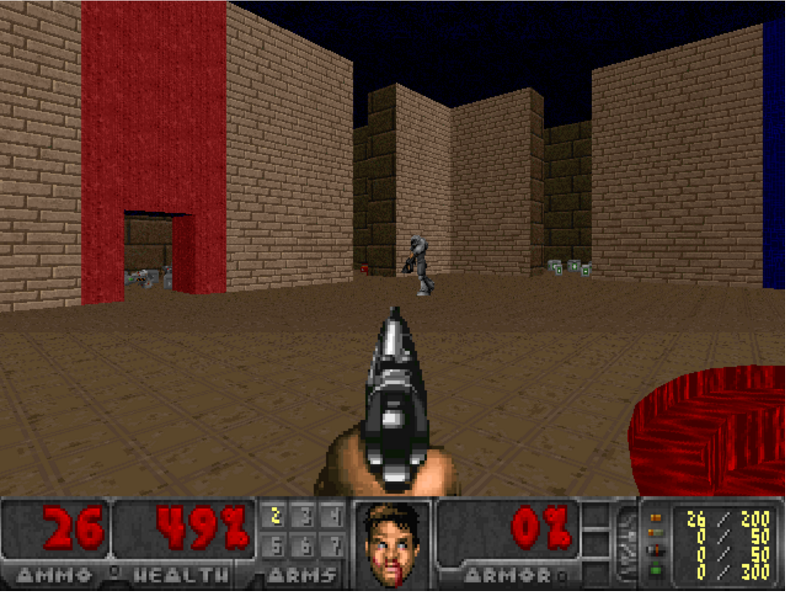}
    &\includegraphics[width=0.32\linewidth]{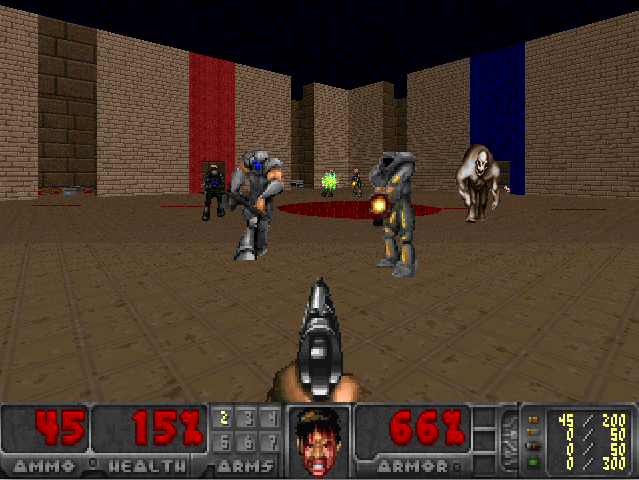}
    &\includegraphics[width=0.32\linewidth]{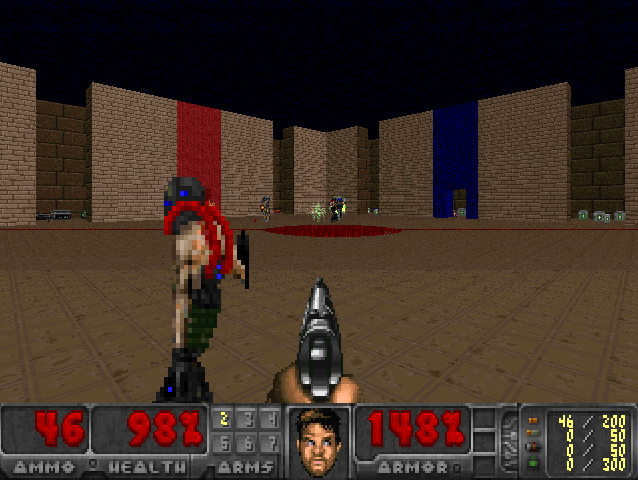}\\[2pt]
    \includegraphics[width=0.32\linewidth]{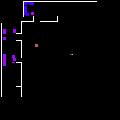}
    &\includegraphics[width=0.32\linewidth]{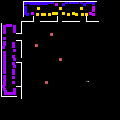}
    &\includegraphics[width=0.32\linewidth]{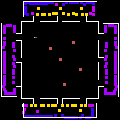}\\[2pt]
  \end{tabular}
  \caption{%
    Motivation:
    As the agent explores the environment, the first-person-view (top)
    only sees a restricted portion of the scene, whereas in the semantic map
    (bottom), the effect of exploration is cumulative, indicating both
    \emph{type} and \emph{position}.
  }
  \label{fig:intro}
\end{figure}

3D domains exhibit a multitude of challenges that cause the standard approaches
discussed above to struggle.
The introduction of an additional spatial dimension, first introduces notions of
partial observability and occlusions, and secondly causes complications due to
viewpoint variance.
Not only is the agent viewing a relatively smaller portion (volume) of the
environment, it also must reconcile observing a variety of other objects in
different contexts under projective transformations.
Furthermore, adding an extra dimension also \emph{combinatorially} complicates
matters in terms of exploration of the environment.
This typically manifests itself in the form of \defoccur{sparse feedback} in
the learning process because the agent's inability to explore the environment
directly penalizes its learning capacity.
Moreover, complications in exploration directly affect any planning that may be
required for tasks and actions.
Finally, with larger search and state spaces comes the likelihood that any
rewards that might help move learning along are also harder to come by.

Sutton \etal (1999) \cite{sutton1999between} propose an extension of the RL
framework that can potentially learn hierarchical policies.
However this, and similar methods, have not been able to scale beyond small
gridworld domains \cite{barto2003recent}.
Kulkarni \etal (2016) \cite{kulkarni2016hierarchical} proposes to tackle
environments with delayed rewards by coupling \defoccur{options} learning and
intrinsically driven exploration methods.
Options are however notoriously hard to train, requiring a great deal of effort
before intrinsically motivated agents can safely deal with generic hierarchical
spatial domains.

Prior work in behavioural modelling and cognitive neuroscience suggests that
humans employ particular, highly specialised mechanisms to construct
representations of, and reason about, the world.
These typically take the form of semantic concepts and abstractions such as
object identity, categories, and localisation.
Freedman and Miller (2008) \cite{freedman2008neural} review evidence from
neurophysiology that explore the learning and representation of object
categories.
Burgess (2008) \cite{burgess2008spatial} discusses evidence from neuroscience for
the presence and combination of different viewpoints (\eg egocentric) and the
role of representing layouts (\eg boundaries and landmarks) in the spatial
cognition process.
Moser \etal (2008) \cite{moser-kropff-moser} also discuss the presence of
highly specialised representation regions in the brain that encode localisation
and spatial reasoning.
Denis and Loomis (2007) \cite{denis2007perspectives} provide a review from
behavioural psychology on the subject of spatial cognition and related topics.

In this paper, we take inspiration from such work to propose a system that
explicitly constructs a joint semantic and topological representation of the
state, and further augment its input with this representation
(Fig.~\ref{fig:intro}) in an attempt to learn policies more effectively in such
complex 3D domains.
To this end, we construct a novel model that incorporates an automatic,
on-the-fly scene reconstruction component into a standard deep-reinforcement
learning framework.
Our work provides a streamlined system to immediately enhance current
state-of-the-art learning algorithms in 3D spatial domains, additionally
obtaining insight on the efficacy of spatially enhanced representations against
those learned in a purely bottom-up manner.
%

%%% Local Variables:
%%% mode: latex
%%% TeX-master: "vizdoom"
%%% ispell-local-dictionary: "british"
%%% End:

%  LocalWords:  approximators Tolman deepRL DQN i RL gridworld etc i

\section{Related work}
\label{sec:related}
\subsection{Deep Reinforcement Learning}
Reinforcement Learning is a commonly employed set of techniques for learning
agents that can execute generic and interactive decision making.
Its mathematical framework is based on \defoccur{Markov Decision Processes}
(MDPs).
An MPD is a tuple $(S, A, P, \reward, \gamma)$, where \(S\) is the set of
states, \(A\) is the set of actions the agent can take at each time step $t$,
$P$ is the transition probability of going from state $s$ to $s'$ using action
$a$, \reward\ is the reward function defining the signal the agent receives
after taking actions and changing states, and $\gamma$ is a discount factor.
The goal of Reinforcement Learning is to learn a policy $\pi : s \rightarrow a$
that maximises the expected discounted average reward over the agent run.
A commonly used technique to learn such a policy is to learn the action-value
function $Q^{\pi}(s, a)$ iteratively, so as to gradually approximate the
expected reward in a model-free fashion.

They have, however, traditionally struggled to deal with high-dimensional
environments, due in large part to the curse of dimensionality.
Deep Reinforcement Learning algorithms such as Deep-Q Networks extend model-free
RL algorithms like Q-Learning to use Deep Neural Networks as function
approximators, implicitly capturing hierarchies in the state representation that
make the RL problem scale even to visual input states.
Unfortunately, they still suffer from some of the problems that standard RL
cannot deal with:
\begin{compactitem}
\item delayed reward signals require non-stochastic exploration strategies
  \cite{kulkarni2016hierarchical};
\item learning to abstract policies hierarchically is currently an unsolved but
  key problem to make RL scale to tasks requiring long-term
  planning \cite{barto2003recent};
\item partial observability in state requires use of models that can encode at
  least short-term memory, specially when training end-to-end
  \cite{hausknecht2015deep}.
\end{compactitem}

Some recent work has also explored ways to develop agents that can learn to play
Doom. Lample and Chaplot (2016) \cite{lample2016playing} take the approach of
using a variant of DRQN \cite{recurrent} together with some game features
extracted directly from the game environment through its data structures.
Our method is similar in the spirit, but can be applied to any environment with
a significant 3D navigation component, as our SLAM and object recognition
pipeline is not intrinsically dependent to the VizDoom platform.
Another interesting approach is the one presented by Dosovitskiy and Koltun
(2016) \cite{doso2016}.
This approach, featured as the winner in the VizDoom
competition~\cite{vizdoom-competition}, changed the supervision signal from a
single scalar reward to a vector of measurements provided by the game engine.
This is used to train a network that, given the visual input, the current
measurements, and the goal, predicts future measurements.
The action to perform is then chosen greedily according the predicted future
measurements.
This is orthogonal to our approach; both algorithms could benefit from 
the novelties introduced by the other, however we leave such
an extension on our part for the future.

\subsection{Simultaneous Localization and Mapping}

Early approaches to camera-pose estimation relied on matching a limited number
of sparse feature points between consecutive video frames \cite{monoslam}.
A common drawback of such solutions is relatively quick error accumulation which
results in significant camera drift.
This may be addressed with PTAM \cite{ptam}, which achieved globally optimal
solutions at real-time rates by running bundle adjustment in a background
thread.
Further improvements include re-localization, loop-closure detection, and faster
matching with binary features \cite{Cummins2009,orbslam,g2o}.
Recently, matching hand-crafted features has been replaced by semi-dense direct
methods \cite{engel14eccv}.
However, these approaches only provide very limited information about the
environment.

A more complete map representation is provided by dense approaches
\cite{Stuhmer2010,dtam,liwicki2016eccv}.
They estimate dense depth from monocular camera input, but their use of a
regular voxel grid limits reconstruction to small volumes due to memory
requirements.
KinectFusion-based approaches \cite{kinectfusion} sense depth measurements
directly using active sensors and fuse them over time to recover high-quality
surfaces, but they too suffer from the same issue.
This drawback has since been removed by scalable approaches that allocate space
only for those voxels that fall within a small distance of the perceived
surfaces, to avoid storing unnecessary data for free space
\cite{niessner2013hashing}.
All approaches mentioned above assume the observed scenes are static.
This assumption can be relaxed by full SLAM with generalized (moving) objects or
some of its more efficient variants \cite{slammot,bibby}, but this is beyond the
scope of this paper.

Approaches such as RatSLAM \cite{milford2004ratslam} and its derivatives propose
instances of SLAM based on models inspired by biological agents, showing
promising results in environments where the navigation task requires some
reasoning about landmarks and non-Cartesian grid representations.

\subsection{Object detection}
Early approaches to object detection include constellation models and pictorial
structures \cite{pictorial_structures}.
The very first object detector capable of real-time detection rates was
\cite{viola_jones}, who solved an inherent problem of sliding-window
approaches by learning a sequential decision process that rapidly rejects
locations which are unlikely to contain any objects.
This concept has since then evolved into a distinct set of algorithms called
\defoccur{proposals}, whose only goal is to quickly localize potential objects
\cite{proposals}.
These locations are then fed into more complex classifiers to determine the
class label (or assign a background).
Deformable part models \cite{dpm} are a prominent example of such, being able to
represent highly variable object classes.
Recently, it has been shown that deformable part models can be interpreted as a
convolutional network \cite{girshick2015dpdpm}, which led to replacement of
handcrafted features by convolutional feature maps \cite{fast-rcnn}.
Finally the Faster-RCNN \cite{renNIPS15fasterrcnn} combines the region proposals
and object detector into a single unified network trainable end-to-end with
shared convolutional features which leads to very fast detection rates.

%%% Local Variables:
%%% mode: latex
%%% TeX-master: "vizdoom"
%%% End:

%  LocalWords:  MDPs MPD maximises iteratively dimensionality RL approximators
%  LocalWords:  observability PTAM voxel KinectFusion scalable voxels RCNN
%  LocalWords:  Deformable deformable convolutional

\section{Semantic Mapping}
\label{sec:mapping}

\begin{figure*}[t]
  \centering
  \includegraphics[width=0.9\textwidth]{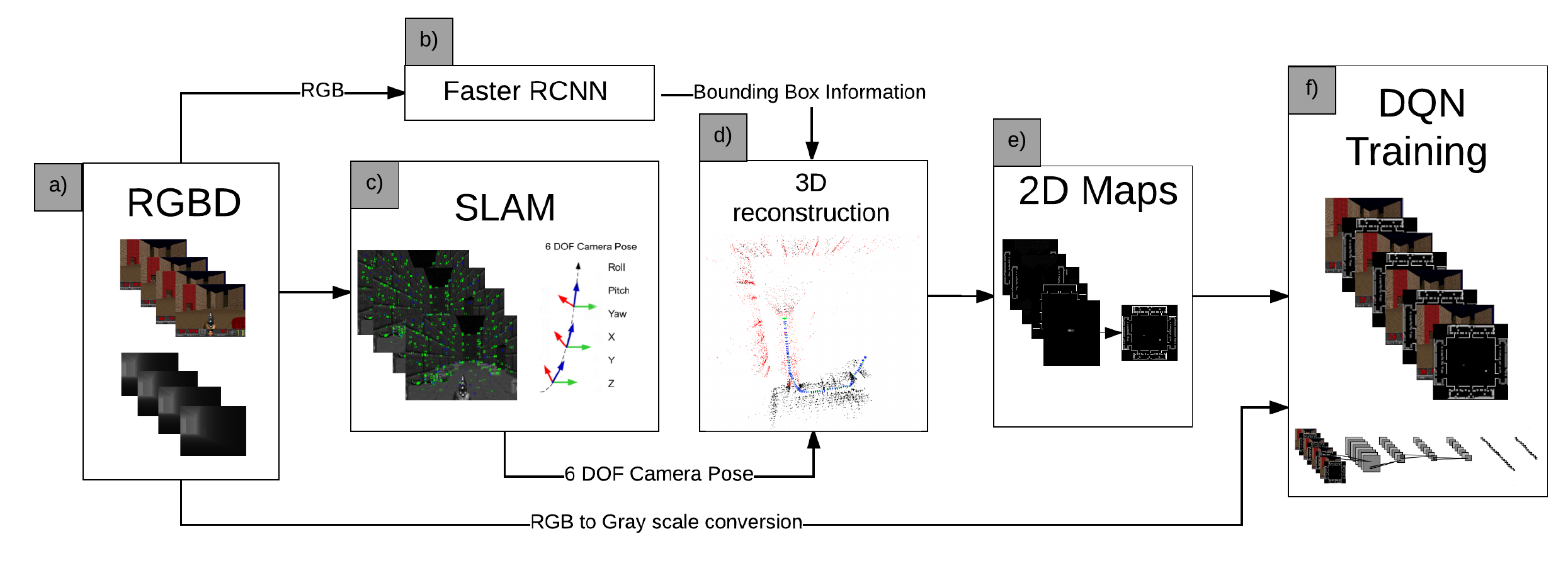}
  \caption{ System overview: (a) Observing image and depth from VizDoom. Running
    Faster-RCNN (b) for object detection and SLAM (c) for pose estimation. Doing
    the 3D reconstruction (d) using the pose and bounding boxes. Semantic maps
    are built (e) from projection and the DQN is trained (f) using these new
    inputs.}
  \label{fig:pipeline_overview}
\end{figure*}

In this paper, we introduce an algorithm based on the Deep Q Network (DQN) that
has been successfully applied to many Atari games \cite{mnih2015human}.
Inspired by prior work in human cognition that indicates how humans employ a
variety of semantic concepts and abstractions (object categories, localization,
\etc) to reason about the world, we build an agent-model that incorporates such
abstractions into its policy-learning framework.
We augment the first-person raw image input to a DQN by adding details about
objects and structural elements encountered, along with the agent’s localization
to cope with complex 3D environments.
This is represented as a 2D map (top-down view) encoding three distinct sources
of information:
\begin{inparaenum}[i)]
\item positions of static structures and obstacles such as walls,
\item position and orientation of the agent, and
\item positions and class labels of important objects such as health packs,
  weapons and enemies.
\end{inparaenum}
Our representation is being updated over time as the agent explores the
environment.
This allows the agent to keep information about areas observed in the past and
build an aggregated model of the 3D environment, as indicated in
Fig.~\ref{fig:intro}.
Such representation allows the agent to behave properly even with respect to the
elements no longer present in the first-person view.

\paragraph{Semantic representation.}%
As the agent explores the environment, we simultaneously estimate localization
of the agent and obstacles (\eg walls) in order to build the map of the
surrounding 3D environment from the first-person-view at each frame.
In parallel, we detect important objects in the scene such as weapons and
ammunition.
And since we want to minimize the dimensionality of the augmented representation
to allow more efficient learning, we project all semantic information onto a
single common 2D map of a fixed size.
Essentially a ``floor-plan'' with the positions of objects and agents.
This is achieved by encoding different entities by different gray-scale values,
in the form of heat-maps (\cf bottom right of Fig.~\ref{fig:intro}).

Our representation encodes position of walls and obstacles (white) extracted
directly from the depth data provided by the VizDoom API.
Information about agent's position and orientation on the 2D map is represented
as a green directed arrow.
We also want to provide the agent semantic information about a variety of
objects present in the environment.
For Doom, we encode the following five object categories: monsters (red), health
packs (purple), high-grade weapons (violet), high-grade ammunition (blue), other
weapons and other ammunition (yellow).
Since these objects could either move or be picked up by another player (\eg\
deathmatch scenario), we project only those objects that are visible in the
current view onto the common map.
This could be addressed by more advanced data association techniques such as
\cite{slammot,bibby}, but this is beyond the scope of this paper.

%%% Local Variables:
%%% mode: latex
%%% TeX-master: "vizdoom"
%%% End:

%  LocalWords:  DQN 3D 2D dimensionality VizDoom RGB deathmatch colour 3D 2D i

\section{Recognition and Reconstruction}
\label{sec:pipeline}
Here, we describe the process of automatically creating semantic maps
on-the-fly.
Fig.~\ref{fig:pipeline_overview} depicts the architecture of our pipeline.
As input, we use the image data provided by the VizDoom API, \ie RGB video
frames visualizing the 3D environment from agents (first person) perspective and
a z-buffer providing depth information of the observed scene.
In order to build a map of the 3D environment, we need to detect and remove all
objects from the z-buffer since we want to
\begin{inparaenum}[i)]
\item provide explicit semantic information about various objects (monsters,
  weapons, \etc) and
\item avoid nuisance visual events such as weapon discharges in the depth
  buffer.
\end{inparaenum}
We also need to know the current pose of the camera, so we run a camera-pose
tracker in parallel with the object detector.
Then, we project the observed scene on a common 3D map and provide its 2D
visualization (top-down view) to the agent.
Note, that the mapping system could work even without access to the z-buffer,
\ie using solely the RGB data \cite{eigen2014depth}.
We now describe the components of our pipeline (object detection, camera pose
estimation and map fusion) in greater detail.

\subsection{Object detection}
\label{subsec:frcnn}
To detect the objects, we use the Faster-RCNN object detector
\cite{renNIPS15fasterrcnn}, which is a convolutional network that combines the
attention mechanism (region proposals) and object detector into a single unified
network, trainable end-to-end.
The first module is a deep fully-convolutional network that simultaneously
predicts object bounds and objectness scores at each position, and the second
module is the Fast R-CNN detector \cite{fast-rcnn} that uses the proposed
regions.
Since both modules share the same features, it offers very fast detection rates.

As input, we use the RGB image resized to the standard resolution of
$227 \times 227$ pixels.
Next, the image is pushed through the network and a convolutional feature map is
extracted.
We use the model of Zeiler and Fergus (2014) \cite{zeiler_fergus} to extract
these feature maps.
To generate region proposals, this feature map is processed in a sliding-window
manner with two fully-connected layers predicting position of the region
proposal and a binary class label indicating ``objectness''.
For each region proposal, the corresponding (shared) feature maps are fed into 2
fully-connected layers with 2048 units that produce soft-max probabilities over
$K$ object classes (and background) and positions of bounding boxes of the
detected objects.
We trained this object detector on five classes corresponding to objects and
monsters that are projected onto the common map.

\subsection{Camera pose estimation}
\label{subsec:slam}
Despite using ground-truth depth maps provided by the z-buffer, ICP-like
approaches \cite{icp} do not work well in game environments since such
environments lack many geometrical features (they are typically represented as
textured planar surfaces to allow fast rendering).
Hence, we use the sparse feature-based ORB-SLAM2 for 6-DoF camera-pose
estimation \cite{orbslam}.
As input, we use RGB images down-sampled to $320\times240$ pixels and a
z-buffer.

First, we build an eight-level image pyramid with a scale factor~\(s_f = 1.2\).
Then, we extract a set of sparse local features representing corner-like
structures.
For this, we use oriented multi-scale FAST detector \cite{fast} with an
adaptively-chosen threshold to detect a sufficient number of features.
The feature extraction step is biased by bucketing to ensure features are
uniformly distributed across space and scale (at least 5 corners per cell).
A constant-velocity motion model predicting the camera pose is used to constrain
matching onto local search windows.
The extracted features are associated with local binary-patterns (256 bits ORB
\cite{orb}) and matched using a mutual-consistency check.
A robust estimate is performed by RANSAC \cite{ransac} with least-squares refinement on the inliers.

Robustness is further increased by keyframes that reduce drift when the camera
viewpoint does not change significantly.
If tracking is lost, the current frame is converted into a bag-of-words and
queried against the database of keyframe candidates for global re-localization.
The camera is re-localized using the PnP algorithm \cite{pnp} with RANSAC.
Global consistency is achieved by loop-closing pose-graph optimization that
distributes the loop-closing error along the graph in a background thread
\cite{g2o}.

\subsection{Mapping}

Once we have the camera poses and a object-masked depth map, we can project the
current frame on a common 3D map.
At each frame $k$, we back-project all image pixels $i$ into the current camera
reference frame to obtain a vertex map $\bf{V}^k_i$
\begin{equation}
  \mathbf{V}^k_i = d^k_i \bf{K}^{-1} \dot{\mathbf{u}}_i.
\end{equation}
Here, $\mathbf{K}^{-1}$ denotes the inverse of the camera calibration matrix
(using parameters from the VizDoom configuration file),
$\dot{\mathbf{u}}_i = [u_i, v_i, 1]^\top$ denote image pixels in homogeneous
coordinates, and $d^k_i$ is depth.
We also want to maintain previously-visited areas in memory so we project the
(homogenized) vertex map $\dot{\mathbf{V}}^k_i = [X_i, Y_i, Z_i, 1]^\top$ from
camera to global reference frame as
$\mathbf{V}^g_i = \mathtt{T}_{g,k} \dot{\mathbf{V}}^k_i$, where
$\mathtt{T}_{g,k} = \{\mathtt{R}, \bft | \mathtt{R} \in \mathbb{SO}_3, \bft \in
\bbR^3\}$ is a rigid body transformation mapping the camera coordinate frame at
time $k$ into the global frame $g$.
Since the fixed volumetric 3D representation severely limits the reconstruction
size that can be handled, we use the hash-based method of
\cite{niessner2013hashing}.

The resulting 2D map is generated by placing a virtual camera at the top-down
view, ignoring all points above and below some height thresholds to remove areas
that would otherwise occlude the map, such as ceilings and floors.

%%% Local Variables:
%%% mode: latex
%%% TeX-master: "vizdoom"
%%% End:

%  LocalWords:  3D VizDoom RGB i ZeroMQ RCNN objectness ICP DoF RANSAC keyframe
%  LocalWords:  keyframes PnP i Vizdoom FRCNN RGBD DQN

\section{Experiments}
\label{sec:experiments}
In this section, we demonstrate the advantage of adding the semantic map
presented in Sec.~\ref{sec:mapping} to the standard first-person view while
working inside the ``Doom'' environment.
Code and results for these experiments will be made available online.
We use the ViZDoom \cite{kempka2016vizdoom} platform for all our experiments.
It is built on top of the first person combat game ``Doom'', and allows easy
synchronous control of the original game, where execution is user-controlled,
getting the first-person-view from the engine at the current step, and stepping
forward by sending it keystrokes.
The environment where the player performs is specified as scenario.

In this paper, we focus on the \defoccur{deathmatch} scenario, in which the map
is a simple arena as can be seen in Fig.~\ref{fig:intro} and the goal is to
eliminate as many opponents as possible before being eliminated.
A proficient agent for this scenario would be the one that is efficient at
eliminating enemies whilst being able to both collect more effective weapons and
keep its own health as high as possible.
This scenario was the basis of the CIG 2016 competition
\cite{vizdoom-competition} where different autonomous agent competed in a
deathmatch tournament.

The quantitative results for all the experiments carried out are summarised in
Tab.~\ref{tab:results}.
The individual features of the experiments run, and the insights obtained from
these runs, are described in subsequent sections, following detailed discussion
of the various components of our framework.
\begin{table}
  \centering
  \begin{tabular}{@{}lD{.}{.}{4}@{}}
    \toprule
    Settings                    & \text{Rewards\hspace*{-3ex}} \\
    \midrule
    Random Play                 & 0.00    \\
    NOSM (player/objects)       & 2.94    \\
    OSM (no FPV)                & 3.16    \\
    baseline                    & 3.45    \\
    NOSM (objects)              & 3.53    \\
    NOSM (walls)                & 3.92    \\
    Prior Dueling DQN           & 5.69    \\
    RSM (localisation)          & 5.87    \\
    OSM (localisation)          & 6.62    \\
    RSM                         & 6.91    \\
    OSM                         & 9.50    \\
    \cmidrule(r){1-2}
    Human Player                & 45.00   \\
    \bottomrule\\[-1pt]
  \end{tabular}
  \caption{%
    Best mean test rewards for the different frameworks run.
    Note that our pipeline performs strongly in comparison to both the
    baselines, and to the ablated versions considered.
    Also note that although the OSM is the best of the artificial systems
    considered, our pipeline, with the RSM is a lot closer to it than the
    others.
  }
  \label{tab:results}
  \vspace*{-3ex}
\end{table}

\paragraph{Recognition and Reconstruction.}%
As described in Sec.~\ref{subsec:frcnn}, we use the Faster-RCNN detector and
feed it with the RGB image given by the platform.
We use a network pre-trained on Imagenet \cite{ILSVRC15} that we fine-tuned on a
dataset consisting of 2000 training and 1000 validation examples extracted from
the ViZDoom engine, performing 5-fold cross-validation.
These images were manually annotated with ground-truth bounding boxes
corresponding to 7 classes: monsters, health packs, high-grade weapons,
high-grade ammunition, other weapons/ammunition, monsters' ammunition, and
agent's ammunition.
After fine-tuning, the model achieved an average precision of $93.21\%$.
The reconstruction system presented in Sec.~\ref{subsec:slam} uses the RGB-D
images provided by the VizDoom platform.

\paragraph{Policy Learning.}%
We use the DQN framework from \cite{mnih2015human} to perform policy learning
with our augmented features.
The only modification to the original algorithm is the CNN architecture that
needs to be able to cope with the extended state.
The first person view (FPV) images are resized to $84\times84$ pixel, converted
to grayscale and normalized.
The semantic 2D map is represented as a single channel image of the same
resolution.
The different object categories are encoded by different grayscale values.
For the experiments that use both the FPV and the 2D map, we concatenate them
along the channel dimension.
The Q network is composed of 3 convolutional layers having respectively, 32, 64
and 64 output channels with filters of sizes $8\times8$, $4\times4$ and
$3\times3$ and 4, 2 and 1 strides.
The fully-connected layer has 512 units and is followed by an output SoftMax
layer. All hidden layers are followed by rectified linear units (ReLU).
Adding the 2D map associated to each FPV image changes input channels from 4 to
8 for the first convolutional layer, and thus increase the number of parameters
from $77824$ to $86016$, a $10\%$ increase.
For training, we use the hyper-parameters from \cite{mnih-atari-2013} and
RMSProp for all experiments.

\paragraph{Action Space.}%
The action space for this environment is an order of magnitude larger than the
Atari environment.
Indeed, ``Doom'' accepts any combination of $43$ unique keystrokes as input.
Following the observation that a human player uses only a small subset of these
combinations to play the game, we recorded actions performed by humans and
selected a representative subset.
These actions can be divided into three groups:
i) actions corresponding to a single keystroke allowing the agent to move and shoot,
ii) combinations of two keystrokes corresponding to moving and shooting at the same time
and iii) actions associated with switching weapons.
We arbitrarily chose the top 13 actions performed by humans, categorising them
into the 3 groups mentioned above.
We did so primarily to constrain the action space to a reasonably tractable
size, while still maintaining richness of actions that could be performed in the
environment.

\paragraph{Reward Function.}%
Our reward function is designed to capture the primary goal of the agent: to
eliminate opponents.
We represent this as $\Delta_{k}$, an indicator variable for an opponent
being eliminated since the last step.
To encourage the agent to live longer, we also consider $\Delta_{h}$, the health
variation between the current step and the previous step.
We explicitly structure the health reward to be zero-sum in order to remove any
biases towards preserving health to the detriment of the primary goal.
The reward~\reward{} incorporating both these terms is written as:
\(\reward = \Delta_{h}/100 + \Delta_{k}\)
where
\(\Delta_{h} \in [-100; 100]\) and \(\Delta_{k} \in \{0,1\}\)

\paragraph{Evaluation Metrics.}%
We use two different scores to evaluate and compare different architectures.
The main metric is the reward function as it allows observing the agent's
behaviour with respect to the primary objective.
The second reported metric is the number of steps the agent has lived.
This is important as living increases the agent's chance to kill opponents and
increase its reward in the longer term.
All reported metrics are mean values over $100$ test games.

\paragraph{Time Complexity.}%
The complete framework has to be fast enough to allow playing at the game's
native speed.
To do so, we run the object detector in parallel with the camera-pose
estimation.
On average, the detector requires $60$ms to process an image while camera-pose
estimation and latency take \(12\)ms and \(10\)ms respectively.
Semantic map construction takes \(25\)ms, and DQN training requires $18$ms to
process a frame and perform one learning step.
The complete pipeline is able to process, on average, 10 images per second.
Given that inside the ViZDoom platform each step represents 4 frames of the game
(as does the Atari emulator), our system plays at approximately 40 frames per
second, which exceeds typical demands of gameplay.
All experiments were run on a Intel Core i7-5930K machine with 32GB RAM and one
NVidia Titan X GPU.

\begin{figure}[h]
	\centering
  \begin{tabular}{@{}c@{}}
   \includegraphics[width=0.4\textwidth]{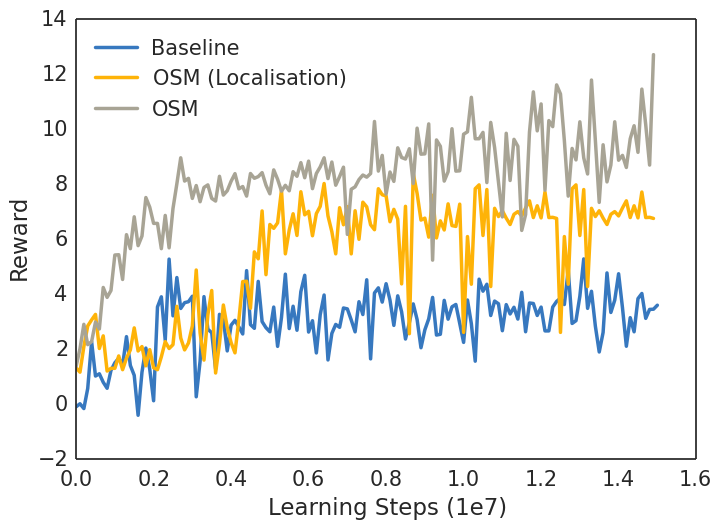}\\[-2pt]
    \includegraphics[width=0.4\textwidth]{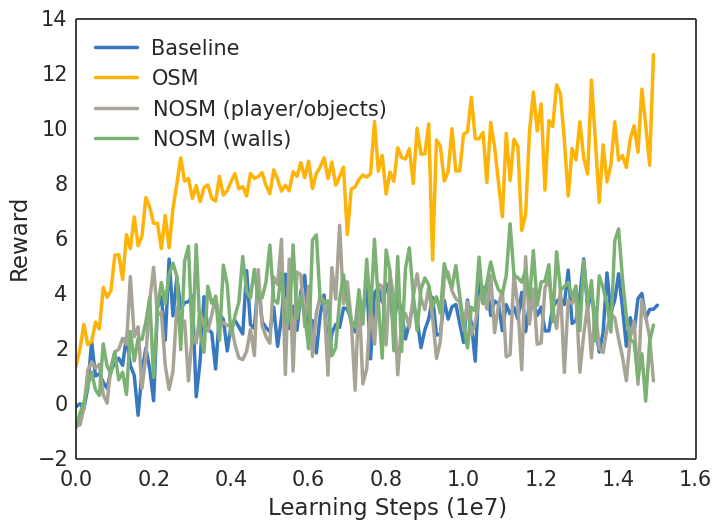}
  \end{tabular}
	\caption{(top) Average reward. (bottom) OSM vs. NOSM.}
  \label{fig:gtmap}
  \vspace*{-3ex}
\end{figure}
\subsection{Oracle Semantic Maps (OSM)}
\label{sec:oracle}
The first set of experiments allows us to evaluate the efficacy of our semantic
representation.
We first isolate potential errors introduced by the recognition and
reconstruction pipeline by extracting ground-truth information about classes and
positions of all objects that are used in the semantic map representation.
In other words, this experiment presents the results we would get if we had
perfect detection and reconstruction, and is used as an ``oracle''.

As the baseline, we use the standard DQN approach trained solely on the first
person view images (referred to as baseline in the following).
This baseline is compared to
\begin{inparaenum}[i)]
\item model trained with both, the first person view and the 2D map encoding ground-truth walls and player position (localisation OSM)
\item model trained with both, the first person view augmented by the complete
  2D maps containing ground-truth walls and positions of player and objects.
\end{inparaenum}

As can be seen in Fig.~\ref{fig:gtmap}, the baseline is not able to learn as good policy as model with our semantic maps.
Moreover, we see that the baseline model quickly reaches a plateau and does not
improve afterwards.
Adding a 2D map of the environment (\ie without objects) allows the agent to
learn a significantly better policy as the reward is almost doubled compared to
the baseline.
Adding the objects seen by the agent onto this map gives another significant
improvement leading to reward of $10$ compared to the $3-4$ achieved by the
baseline.
Moreover, we can see that the network provided with the complete 2D map
(including objects) is able to learn faster than the models provided with fewer
information.
This result proves that providing higher level, complex representation of the
surrounding of the agent allows it to learn faster and converge to a better
policy.

\begin{figure}[h]
  \centering
  \begin{tabular}{@{}c@{\hspace*{5pt}}c@{\hspace*{5pt}}c@{}}
    \begin{tabular}{@{}c@{}}
        \includegraphics[width=0.315\linewidth]{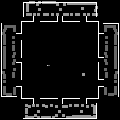}\\%[-1ex]
        \includegraphics[width=0.315\linewidth]{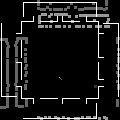}
    \end{tabular}
 &
   \begin{tabular}{@{}c@{}}
        \includegraphics[width=0.315\linewidth]{graphics/final}\\%[-1ex]
        \includegraphics[width=0.315\linewidth]{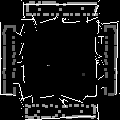}
   \end{tabular}
 &
   \begin{tabular}{@{}c@{}}
        \includegraphics[width=0.335\linewidth]{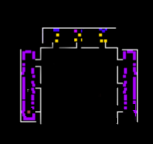}\\%[-1ex]
        \includegraphics[width=0.335\linewidth]{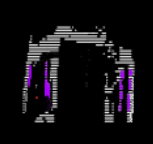}
   \end{tabular}
  \end{tabular}
  \caption{%
    The top maps for each column are all taken from the oracle.
    The maps on the bottom are
    (l) Oracle map with noise on player and objects' positions.
    (m) Oracle map with noise on the walls.
    (r) Semantic map reconstructed, independent of the oracle, by our pipeline.
  }
  \label{fig:some-maps}
\end{figure}
\subsection{Noisy Oracle Semantic Maps (NOSM)}
Unfortunately, the detection and reconstruction pipelines are often imperfect in
real world scenarios.
Next, we study the impact of providing a very poor spatial representation to the
agent.
To do that, we add a significant amount of noise to the ground-truth data
extracted from the game to see how the DQN framework reacts.

First, we consider the case where we add the same Gaussian noise to the agent's
and all objects' positions, referenced as NOSM (player/objects), meaning that
these elements are not properly positioned with respect to the static objects.
Fig.~\ref{fig:some-maps}(l) shows the results of adding that noise.
The OSM map is shown on top and its noisy version is shown below.
One thing to note here is that these maps have gray scale pixel values to define
different abstractions and objects.
This gray scaled format was used for training as discussed in the previous
sections.
Next, we add Gaussian noise to the positions of walls, referenced as NOSM
(walls), meaning that some element that appear accessible in the 2D map cannot
be reached in the real environment.
Fig.~\ref{fig:some-maps}(m) shows the results of adding that noise.

As can be seen in Fig.~\ref{fig:gtmap}(bottom), this very high amount of noise
in the 2D maps prevent the DQN framework to learn a good policy.
However, it is important to note that in the worst case, the noisy version
matches the performances of the baseline as the network learns to ignore it.

\subsection{Reconstructed Semantic Maps (RSM)}

In Sec.~\ref{sec:oracle}, we have shown the efficacy of Q-learning with ground-truth version of our semantic maps.
As a proof of concept, we now evaluate performance with the real maps generated
on-the-fly by the approach described in Sec.~\ref{sec:pipeline} (RSM).
This experiment allows us to evaluate the quality of the policy that can be
learned when using the standard detection and mapping techniques without any
extra engineering.
In other words, we measure the drop in performance caused by imperfect object
detection and SLAM in a real world scenario with respect to the oracle.
The difference between the OSM and the RSM is seen in
Fig.~\ref{fig:some-maps}(r).
Here, the semantic categories are coloured instead of greyscale levels for
emphasis.

\begin{figure*}[t]
  \centering
  \begin{tabular}{@{}c@{\hspace*{2pt}}c@{\hspace*{2pt}}c@{}}
    \includegraphics[width=0.33\textwidth]{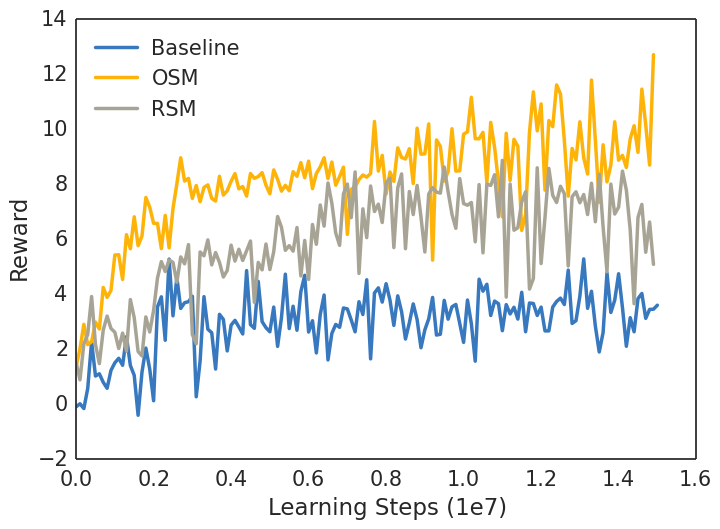}
    &\includegraphics[width=0.33\textwidth]{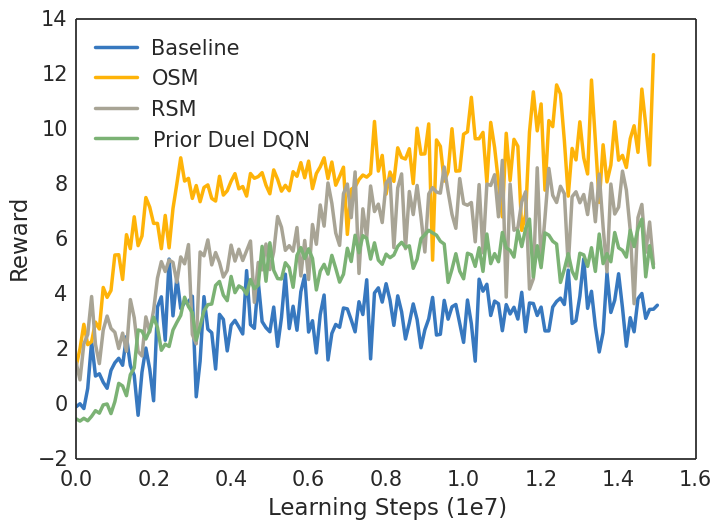}
    &\includegraphics[width=0.33\linewidth]{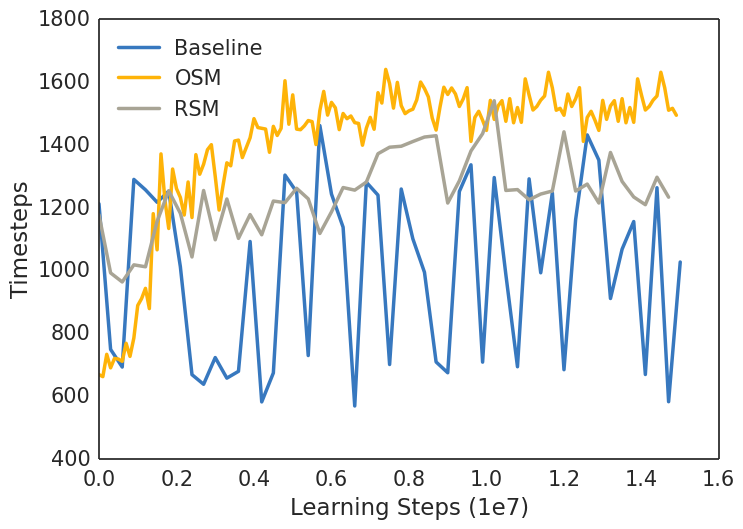}\\
  \end{tabular}
  \caption{(l) OSM vs. RSM (m) Our method vs. dual
    DQN with prioritized ER. (r) OSM vs. DQN on mean run-length}
  \label{fig:graphs}
  \vspace{-0.25cm}
\end{figure*}

As seen in Fig.~\ref{fig:graphs}(l), the reconstructed map leads to
significantly better results than the baseline.
Even though it doesn't match the oracle, we clearly see that the RSM is much
closer to the OSM than the baseline.
The remaining gap can be further reduced with progress in the field.

\subsection{Prioritized Duel DQN}
Combination of the prioritized experience replay \cite{prioritized} and dueling
network architecture \cite{duel} has demonstrated superior results on 57 Atari
games (2D environment) compared to the vanilla DQN approach that is the baseline
considered above.
In this experiment, we compare this successful model (referred as dDQN) with the basic DQN model augmented with our semantic maps.

Fig.~\ref{fig:graphs}(m) shows that while the combination of PRL with dual DQN
achieves better results than the DQN baseline, the model with our semantic maps,
despite trained with the basic DQN, outperformed the PRL with dual DQN trained
on first person views.
It is also interesting to note that these two approaches are orthogonal and could be combined.
We leave this study for the future work.

\subsection{Mean Run Length}
As can be seen in Fig.~\ref{fig:graphs}(r), the agent trained with semantic maps
is able to typically live longer than the one trained only on the first-person
view.
This is a consequence of the fact that the OSM agent inherently attempts to
build a representation of the environment it is in, which helps it adapt better
from arbitrary initialisation points.
The baseline however, does not have access to such capabilities, and hence
performs incoherently in these situations.
In keeping with the general characteristics of the results seen thus far, the
RMS agent typically underperforms in relation to the ORM agent, but still
significantly outperforms the baseline.

%%% Local Variables:
%%% mode: latex
%%% TeX-master: "vizdoom"
%%% End:

%  LocalWords:  subsec frcnn RCNN Imagenet ViZDoom VizDoom DQN FPV grayscale i
%  LocalWords:  SoftMax ReLU RMSProp OSM NOSM gtmap center doesn dueling dDQN
%  LocalWords:  PRL SRM ORM

\section{Discussion and Conclusion}
\label{sec:discussion}

We proposed to augment the standard DQN model with semantic maps; a
representation that provides aggregated information about the 3D environment
around the agent.
We have demonstrated the efficacy of our approach with both oracle maps, and
automatically reconstructed maps using object detection and SLAM, demonstrating
the efficacy of our approach with standard computer-vision recognition and
reconstruction pipeline (\eg for road scene understanding~\cite{vineet2015icra})
and a standard off-the-shelf policy learner (DQN).

Our central thesis is exploring the benefits of semantic representations
augmenting the directly-from-pixels learning approach typically employed.
While we do not claim major contributions to policy-learning algorithms
themselves, the effort nonetheless provides insight on the efficacy of such
representations against those learned in a purely bottom-up manner.
It also potentially serves as a benchmark for effectiveness of representations
learned in a purely bottom-up manner.
Moreover, our approach has the potential to extend and scale beyond the Doom
environment by virtue of its applicability to any environment with a reasonable
number of potential other entities and the extractability of 3D information.

In terms of future directions, we would like to extend our framework
along a variety of different axes.
One particular direction is improving our resilience to layered environments as
we are currently unable to represent environments such as buildings (``stacked
floors/levels'').
Another direction involves relaxing the metric constraints that our maps are
currently constructed under.
Better localisation and semantic representations could exist that do not
necessarily require metric reconstruction, but perhaps a more relativistic,
graph-based approach.
And finally, we are interested in extending our experiments to incorporate more
maps (beyond the deathmatch scenario we currently employ), and elicit
qualitative judgments of the learned gameplay.

%%% Local Variables:
%%% mode: latex
%%% TeX-master: "vizdoom"
%%% End:

%  LocalWords:  DQN

%\clearpage
{\small
\bibliographystyle{ieee}
\bibliography{references}
}

\end{document}